\documentclass[12pt]{article}
\usepackage{bbm}
\usepackage{amsfonts}
\usepackage{amsmath,amssymb,pstricks,fullpage}
\usepackage{subfigure}
\usepackage{graphicx}
\usepackage{natbib}
\usepackage{jstyle}
\usepackage{verbatim}
\usepackage{algorithm}

\def\ds{\displaystyle}
\let\hat\widehat
\let\tilde\widetilde

\def\E{\mathbb{E}}
\def\P{\mathbb{P}}

\def\Var{\text{Var}}
\def\reals{\mathbb{R}}

\def\qed{\hskip1pt $\;\;\scriptstyle\Box$}
\def\c3po{$\text{C3PO}$}

\def\P{{\mathbb P}}
\def\E{{\mathbb E}}
\def\reals{{\mathbb R}}

\def\ds{\displaystyle}
\def\argmin{\mathop{\text{arg\,min}}}

\def\Var{\textrm{Var}}
\newcommand{\beq}{\begin{eqnarray*}}
\newcommand{\eeq}{\end{eqnarray*}}
\newcommand{\beqn}{\begin{eqnarray}}
\newcommand{\eeqn}{\end{eqnarray}}

\def\supp{{\text{supp}}}

\def\Sc{S^c}

\def\bm#1{{#1}}

\let\hat\widehat
\let\tilde\widetilde
\let\hat\widehat

\def\P{{\mathbb P}}
\def\E{{\mathbb E}}
\def\reals{{\mathbb R}}

\def\ds{\displaystyle}
\def\argmin{\mathop{\text{arg\,min}}}

\newtheorem{assumption}{Assumption}

\title{ \Huge  On the $\ell_1\text{-}\ell_q$ Regularized Regression}

\author{Han Liu$^{1,2}$, Jian Zhang$^3$ \\
$^1$Machine Learning Department, $^2$Statistics Department\\
  Carnegie Mellon University, Pittsburgh, PA, 15213 \\
  $^3$Department of Statistics \\
  Purdue University, West Lafayette, IN, 47907}

\begin{document}
\maketitle

\begin{center}
{\bf ABSTRACT}
\end{center}

\begin{quote}
{\small In this paper we consider the problem of grouped variable
selection in high-dimensional regression using
$\ell_1\text{-}\ell_q$ regularization  ($1\leq q \leq \infty$),
which can be viewed as a natural generalization of the
$\ell_1\text{-}\ell_2$ regularization (the group Lasso). The key
condition is that the dimensionality $p_n$ can increase much faster
than the sample size $n$, i.e.  $p_n \gg n$ (in our case $p_n$ is
the number of groups),  but the number of relevant groups is small.
The main conclusion is that many good properties from
$\ell_1\text{-}$regularization (Lasso) naturally carry on to the
$\ell_1\text{-}\ell_q$ cases ($1 \leq q \leq \infty$), even if the
number of variables within each group also increases with the sample
size. With fixed design, we show that the whole family of estimators
are both estimation consistent and variable selection consistent
under different conditions. We also show the persistency result with
random design under a much weaker condition. These results provide a
unified treatment for the whole family of estimators ranging from
$q=1$ (Lasso) to
 $q=\infty$ (iCAP), with $q=2$ (group Lasso)as a special case.
When there is no group structure available, all the analysis reduces
to the current results of the Lasso estimator ($q=1$).  }
\end{quote}

\vspace{0.3in} {\bf \noindent Keywords:} $\ell_1\text{-}\ell_q$
regularization, $\ell_1\text{-}$consistency, variable selection
consistency, sparsity oracle inequalities, rates of convergence,
Lasso, iCAP, group Lasso, simultaneous Lasso

\section{Introduction}\label{introduction}

We consider the problem of recovering a high-dimensional vector
$\beta^* \in \mathbb{R}^{m_n}$ using a sample of independent pairs
$(X_{1\bullet},Y_1),\ldots, (X_{n\bullet},Y_n)$ from a multiple
linear regression model, $Y =X\beta^* + \epsilon $. Here $Y$ is the
$n \times 1$ response vector and $X$ represents the observed
$n\times m_n$ design matrix whose $i$-th row vector is denoted by
$X_{i\bullet}$. $\beta^*$ is the true unknown coefficient vector
that we want to recover, and $\epsilon = ( \epsilon_1,\ldots,
\epsilon_n)$ is an $n \times 1$ vector of i.i.d. noise with
$\epsilon_i \sim \mathcal{N}(0,\sigma^2)$.

In this paper we are interested in the situation where all the
variables are naturally partitioned into $p_n$ groups. Grouped
variables often appear in real world applications. For example, in
many data mining problems we encode categorical variables using a
set of dummy variables and as a result they form a group. Another
example is additive model, where each component function can be
represented using its basis expansions which can be treated as a
group. Suppose the number of variables in the $j$-th group is
represented by $d_j$, then by definition we have $m_n =
\sum_{j=1}^{p_n}d_j$.
We can rewrite the above linear model as
\begin{eqnarray}
Y = X \beta^* + \epsilon = \sum_{j=1}^{p_n} X_j \beta^*_j + \epsilon
\end{eqnarray}
where $X_j$ is an $n \times d_j$ matrix corresponding to the $j$-th
group (which could be either categorical or continuous) and
$\beta^*_j$ is the corresponding $d_j \times 1$ coefficient
subvector. Therefore, we have $X=(X_1, \ldots, X_{p_n})$ and
$\beta^*=({\beta^*}_1^T, \dots, {\beta^*}_{p_n}^T)^T$. All
predictors and the response variable are assumed to be centered at
zero to simplify notation. Furthermore, we use $X_{\underline{j}}$
to represent the $j$-th column in the design matrix $X$ and assume
that all columns in the design matrix are standardized, i.e. $\ds
\frac{1}{n}\|X_{\underline{j}}\|^2_{\ell_2} = 1, \underline{j} =
1,\ldots, m_n$.  Similar to the notation of $X_{\underline{j}}$, we
denote $\beta^{*}_{\underline{j}}$ ($\underline{j} = 1, \ldots,
m_n$) to be the $j$-th  individual element of the vector $\beta^*$.
Since we are mainly interested in the high-dimensional setting, we
allow the number of groups $p_n$ to increase as the number of
examples $n$ increases and our results mainly focus on the case
where $p_n \gg n$. Furthermore, we also allow the group size $d_j$
to increase with $n$ at a rate $d_j = o(n)$ and define ${\bar{d}_n}
= \max_{j} d_j$ to be the upper bound of the group size for a fixed
$n$. In the rest of the paper we will suppress the subscript $n$
when there is no confusion.

In order to obtain a reliable estimation of $\beta^*$ when $p_n \gg
n$, the key assumption is that the true coefficient vector $\beta^*$
is {\it sparse}. Denote $S = \{j: \|\beta^*_j \|_{\ell_\infty}\neq
0, j = 1,\ldots, p_n\}$ to be the set of group indices and let $s_n
= |S|$ to be the cardinality of the set $S$, we also denote
$\beta^*_S$ to be the vector concatenating all subvectors
$\beta^*_j$'s for $j \in S$. The sparsity assumption means that $s_n
\ll p_n$. Therefore, even if $\beta^*$ has a very high dimension,
the only effective part is $\beta^*_S$ while the remaining part
$\beta^*_{S^c} = \mathbf{0}$. Our task is to select and recover the
nonzero groups of variables corresponding to the index set $S$.

Sparsity has a long history of successes in solving such
high-dimensional problems. Without considering the group structure,
there exist many classical methods for variable selection, such as
AIC \citep{akai:1973}, BIC \citep{Schw:1978}, Mallow's $C_p$
\citep{Mall:1973}, etc. Although these methods have been proven to
be theoretically sound and have been shown to perform well in
practice, they are only computationally feasible when the number of
variables is small. Recently, more attention has been focused on the
$\ell_1\text{-}$regularized least squares (Lasso) estimator
\citep{Tibshirani:96, Chen:Dono:Saun:1998} which is defined as
\begin{eqnarray}
\hat{\beta}^{\lambda_n} = \argmin_\beta \Biggl\{\frac{1}{2n}\|Y -
X\beta\|_{\ell_2}^2 + \lambda_n\|\beta \|_{\ell_1}\Biggr\}
\end{eqnarray}
where $\lambda_n$ is the regularization parameter for the
$\ell_1$-norm of the coefficients $\beta$, while
$\hat{\beta}^{\lambda_n}$ means the Lasso solution when $\lambda_n$
is used for regularization. In the following, we will suppress the
superscript if not confusion is caused. Lasso can be formulated as a
quadratic programming problem and the solution can be solved
efficiently \citep{Osborne:2000, lars}. Its asymptotic properties
for fixed dimensionality have been studied in \citep{Knight:00}. For
high dimensional setting, \cite{GR04} prove that Lasso estimator is
persistent, in the sense that, when constrained in a class, the
predictive risk of the Lasso estimator converges to the risk
obtained by the oracle estimator in probability. However, recent
studies \citep{Meinshausen:2006, ZY07,Zou:2006} show that the Lasso
estimator is not in general variable selection consistent, which
means that in general the correct sparse subset of the relevant
variables can not be identified even asymptotically. In particular,
in \citep{ZY07, Martin:06}, it is shown that in order for Lasso to
be variable selection consistent, the so-called irrepresentable
condition has to be satisfied. \cite{Zou:2006} propose the adaptive
Lasso and show that by using adaptive weights for different
variables, the $\ell_1$ penalty can lead to variable selection
consistency. In terms of estimation, it has been show in \cite{MY06}
that under weaker conditions, the Lasso estimator is
$\ell_2\text{-}$consistent for high-dimensional setting where the
total number of variables can grow almost as fast as $\exp(n)$.
Under a stronger assumption, \cite{Bunea:annal:07} further proves
the sparsity oracle inequalities for the Lasso estimator using fixed
design, which bounds the $\ell_2$-norm of the predictive error in
terms of the number of non-zero components of the oracle vector.
Such results can be used applied to nonparametric adaptive
regression estimation and to the problem of aggregation of arbitrary
estimators. Parallel to the fixed design result, a similar result
for the random design can be found in \citep{Bunea:07}. A more
recent result from \citep{Bick:Ritov:2007} refine similar oracle
inequalities using weaker assumptions. All these results show that
for sparse linear models, Lasso can overcome the curse of
dimensionality even when facing increasing dimensions.

When variables are naturally grouped together, it is more meaningful
to select variables at a group level instead of individual
variables, as can be seen from previous examples.
A general strategy for grouped variable selection is to use block
$\ell_1\text{-}$norm regularization. For variables within each block
(group), an $\ell_q$ norm is applied, and different blocks are then
combined by an $\ell_1$ norm (therefore the name
$\ell_1\text{-}\ell_q$ regularization). One such example is the
group Lasso \citep{Yuan:glasso:06}, which is an extension of Lasso
for grouped variable and can be viewed an $\ell_1\text{-}\ell_2$
regularized regression. Other works related to grouped variable
selection include the iCAP estimator~\citep{cap08}, which can be
viewed as an $\ell_1\text{-}\ell_\infty$ regularized regression, and
group logistic regression ~\citep{grlasso:2007}, etc. Using random
design, \cite{grlasso:2007} proved the estimation consistency result
for group Lasso with Lipschitz type loss functions. Also with random
design, \cite{Bach:07} derived a similar irrepresentable condition
as in~\citep{ZY07} and proved the variable selection consistency
result for group Lasso. However, to the best of our knowledge, there
isn't corresponding result for estimation and variable selection
consistency for the group Lasso and iCAP estimators using fixed
design, nor the persistency results using random design.  There is
also no systematic theoretical treatment for the whole family of the
more general $\ell_1\text{-}\ell_q$ regularized regression with $1
\leq q \leq \infty$.

Our work tries to bridge this gap and provide a unified treatment of
$\ell_1\text{-}\ell_q$ regularized regression for the whole range
from $q=1$ to $q=\infty$. The main conclusion of our study is that
many good properties from $\ell_1\text{-}$regularization (Lasso)
naturally carry on to the $\ell_1\text{-}\ell_q$ cases ($1 \leq q
\leq \infty$), even if the number of variables within each group can
increase with the sample size $n$. Using fixed design, when
different conditions are assumed, we show that
$\ell_1\text{-}\ell_q$ estimator is both estimation consistent and
variable selection consistent, and if the linear model assumption
does not hold, sparsity oracle inequalities for the prediction error
could still be obtained under a weaker condition. Using random
design, we show that a constrained form of the
$\ell_1\text{-}\ell_q$ regression estimator is persistent. Our
results provide simultaneous analysis to both the iCAP ($q =
\infty$) and the group Lasso estimators ($q=2$). When there is no
group structure, all the analysis naturally reduces to the current
results of the Lasso estimator ($q=1$). One interesting application
of these results is to analyze the simultaneous Lasso estimator
\citep{Turlach05}, which can be viewed as an
$\ell_1\text{-}\ell_\infty$ regularized regression using block
designs.

The rest of the paper is organized as follows. In Section
\ref{intro} we first introduce  some preliminaries of the
$\ell_1\text{-}\ell_q$ regularized regression and then describe some
characteristics of its solution.  In Section \ref{sparsistency}, we
study the variable selection consistency result. In Section
{\ref{sec.consistency}}, we study the estimation consistency and the
sparsity oracle inequalities. In Section \ref{sec.persistency}, we
study the persistency property. We conclude with some discussion in
Section \ref{sec.conclusion}.

\section{$\ell_1\text{-}\ell_q$ Regularized Regression}
\label{intro}

Given the design matrix $X$ and the response vector $Y$, the
$\ell_1\text{-}\ell_q$ regularized regression estimator is defined
as the solution of the following convex optimization problem:
\begin{equation}\label{eqn.grouplasso}
\hat{\beta}^{\lambda_n} = \mathop{\arg\min}_\beta \frac{1}{2n}\|Y -
X\beta\|^2_{\ell_2} + \lambda_n
\sum_{j=1}^{p_n}(d_j)^{1/q'}\|\beta_j\|_{\ell_q}
\end{equation}
where $\lambda_n$ is a positive number which penalizes complex model
and $q'$ is the conjugate exponent of $q$, which satisfies $\ds
\frac{1}{q'} + \frac{1}{q} = 1$ (assuming $\frac{1}{\infty} = 0$).
The terms $(d_j)^{1/q'}$ are used to adjust the effect of different
group sizes. It is easy to see that when $q=1$, this reduces to the
standard Lasso estimator; when $q=2$, this reduces to the group
Lasso estimator \citep{Yuan:glasso:06}; when $q=\infty$, this
reduces to the $\ell_1\text{-}\ell_\infty$ regularized regression
estimator, or the iCAP estimator defined in~\citep{cap08}.

To characterize the solution to this problem, the following result
can be straightforwardly obtained using the Karush-Kuhn-Tucker (KKT)
optimality condition for convex optimization.

\begin{proposition}\label{proposition.kkt} (KKT conditions)
A vector $\hat{\beta} = (\hat{\beta}_1^T, \ldots,
\hat{\beta}_{p}^T)^T \in \mathbb{R}^{m_n}$, $m_n =
\sum_{j=1}^{p_n}d_j$,  is an optimum of the objective function in
\eqref{eqn.grouplasso} if and only if there exists a sequence of
subgradients $\hat{g}_j \in  \partial \|\hat{\beta}_j \|_{\ell_q}$,
such that
\begin{equation}
\label{eq:dbstat} \frac{1}{n}X^T_j\left(X\hat\beta - Y\right) +
\lambda_{n}(d_j)^{1/q'}\hat{g}_j = \mathbf{0}.
\end{equation}
The subdifferentials $ \partial \|\hat{\beta}_j \|_{\ell_q}$ is the
set of vectors $\hat{g}_j\in \reals^{d_j}$ satisfying

If $1 < q < \infty$, then
\begin{eqnarray}
\ds \hat{g}_j = \partial\|\hat{\beta}_{j}\|_{\ell_q} =
\left\{\begin{array}{cc}
                                  B^{q'}(1) & \mathrm{if}~
                                  \hat{\beta}_j=\mathbf{0} \\
                                 \ds \Biggl\{\biggl( \frac{|\hat{{\beta_j}}_{\ell}|^{q-1}\mathrm{sign}(\hat{{\beta_j}}_{\ell})}{\|\hat{\beta}_j\|^{{q-1}}_{\ell_q}} \biggr)_{\ell=1}^{d_j} \Biggr\}  &
                                  \mathrm{o.w.}
                                \end{array}
 \right.
\end{eqnarray}
where $B^{q'}(1)$ denotes the ball of radius 1 in the dual norm,
i.e. $1/q + 1/q' = 1$. It's easy to see that
$\|\hat{g}_j\|_{\ell_{q'}} \leq 1$ for  any $j$.

If $q=\infty$ then
\begin{eqnarray}
\hat{g}_j =  \partial\|\hat{\beta}_j\|_{\ell_\infty} =
\left\{\begin{array}{cc}
                                               B^1(1) & \mathrm{if}~\hat{\beta}_j=\mathbf{0} \\
                                              \ds \mathrm{conv}\{\mathrm{sign}({\hat{\beta_j}}_\ell)e_\ell:|{\hat{\beta_j}}_\ell| = \|\hat{\beta}_j \|_{\ell_\infty} \}&
                                               \mathrm{o.w.}
                                             \end{array}
  \right.
\end{eqnarray}
where $\mathrm{conv}(A)$ denotes the convex hull of a set A and
$e_\ell$ the $\ell$-th canonical unit vector in $\mathbb{R}^{d_j}$.
It's also easy to see that $\|\hat{g}_j\|_{\ell_{q'}} = \|
\hat{g}_j\|_{\ell_1} \leq 1$ for all $j$ when $q = \infty$.

If $q=1$ then
\begin{eqnarray}
\hat{g}_j =  \partial\|\hat{\beta}_j\|_{\ell_1} = \{\xi \in
\mathbb{R}^{d_j}: \xi_\ell \in \partial|\cdot|(x_\ell),
\ell=1,\ldots, d_j \}.
\end{eqnarray}
\end{proposition}

From proposition \ref{proposition.kkt}, the $\ell_1\text{-}\ell_q$
regularized regression estimator can be efficiently solved even with
large $n$ and $p_n$. For example, blockwise coordinate descent
algorithms as in \citep{cap08} can be easily applied. When $q=1$ and
$q = \infty$, due to fact that feasible parameters are constrained
to lie within a polyhedral region with parallel level curves,
efficient path algorithm can be developed \citep{lars,cap08}. At
each iteration of the blockwise coordinate descent algorithm,
$\beta_j$ for $j=1,\ldots,p_n$ is updated, with the rest of the
coefficients fixed. Coupled with a threshold operator, these
algorithms general converge very fast and exact solution can be
obtained. Standard optimization methods, such as interior-point
methods \citep{boyd:2004}, can also be directly applied to solve the
$\ell_1\text{-}\ell_q$ regularized regression problems.

It is well-known \citep{Osborne:2000} that under some conditions,
the Lasso can at most select $n$ nonzero variables even in the case
$p_n \gg n$. A similar but weaker result can be obtained for the
$\ell_1\text{-}\ell_q$ regularized regression.

\begin{proposition}\label{proposition.glasso.solution}
For the $\ell_1\text{-}\ell_q$ regularized regression problem
defined in equation (\ref{eqn.grouplasso}) with $\lambda_n>0$, there
exists a solution $\hat{\beta}^\lambda$ such that the number of
nonzero groups $|S(\hat{\beta})|$ is upper bounded by $n$, the
number of given data points, where $S(\hat{\beta}) =
\{j:\hat{\beta}_j \ne \mathbf{0}\}$
\end{proposition}

\begin{remark}Notice that the solution to $\ell_1\text{-}\ell_q$ regularized regression problem may not be unique especially when
$p_n \gg n$ (similar to the Lasso case), since the optimization
problem might not be strictly convex. Consequently, there might
exist other solutions that contain more than $n$ active groups.
However, a compact solution $\hat{\beta}$ with $|S(\hat{\beta})| \le
n$ can always be obtained by following an easy and mechanical step
described in the proof of Proposition
\ref{proposition.glasso.solution}.
\end{remark}
\noindent {\bf Proof}: From the KKT condition in proposition
\ref{proposition.kkt}, we know that any solution $\hat{\beta}$
should satisfy the following conditions ($j=1,\ldots,p_n$):
$$
\frac{1}{n}X_j^T (Y - X \hat{\beta}) = \lambda g_j
$$
where $\ds g_j = \partial \|\beta_j \|_{\ell_q}$. Now suppose there
is a solution $\hat{\beta}$ which has $s=|S(\hat{\beta})| > n$
number of active groups, in the following we will show that we can
always construct another solution $\tilde{\beta}$ with one less
active group, i.e. $|S(\tilde{\beta})| = |S(\hat{\beta})| - 1$.

Without loss of generality assume that the first $s$ groups of
variables in $\hat{\beta}$ are active, i.e. $\hat{\beta}_j \ne
\mathbf{0}$ for $j=1,\ldots, s$. Since
$$
X \hat{\beta} = \sum_{j=1}^{s} X_j \hat{\beta}_j \in \mathbb{R}^{n
\times 1}
$$
and $s > n$, the set of vectors $X_1 \hat{\beta}_1, \ldots,
X_{s}\hat{\beta}_{s}$ are linearly dependent. Without loss of
generality assume
$$
X_1 \hat{\beta}_1 = \alpha_2 X_2\hat{\beta}_2 + \ldots + \alpha_s
X_s \hat{\beta}_s.
$$
Now define $\tilde{\beta}_j = \mathbf{0}$ for $j=1$ and $j>s$, and
$\tilde{\beta}_j = (1+\alpha_j)\hat{\beta}_j$ for $j=2,\ldots,s$,
and it is straightforward to check that $\tilde{\beta}$ satisfies
the KKT condition and thus is also a solution to the
$\ell_1\text{-}\ell_q$ regularized regression problem in equation
\ref{eqn.grouplasso}. The result thus follows by induction. $\Box$

The main objective of the paper is to investigate several important
statistical properties of the $\ell_1\text{-}\ell_q$ estimator
$\hat{\beta}$. We first give some rough definitions of the
properties that we would like to establish, more details will be
shown in their corresponding sections.

\begin{definition}
(Variable selection consistency) An estimator is said to be variable
selection consistent if it can correctly recover the sparsity
pattern with probability goes to 1. For the case of grouped variable
selection, $\hat{\beta}$ is said to be variable selection consistent
if
\begin{eqnarray}
\mathbb{P}\left( S(\hat{\beta}) = S(\beta^*)\right) \rightarrow 1.
\end{eqnarray}
\end{definition}

\begin{definition}($\ell_1\text{-}$estimation consistency)  An estimator is said to be $\ell_1\text{-}$estimation
 consistent if the $\ell_1\text{-}$norm of the difference between
 the estimator and the true parameter vector converges to 0 in
 probability. i.e.
 \begin{eqnarray}
\forall \delta >0~~~\mathbb{P}\left(\|\hat{\beta}-\beta^*
\|_{\ell_1}
> \delta \right) \rightarrow 0.
 \end{eqnarray}

\end{definition}

\begin{definition}(Prediction error consistency)  An estimator is said to
be prediction error consistent if the prediction error, defined as
$\ds \frac{1}{n}\|\hat{Y} - X\beta^* \|_{\ell_2}^2$, of the
estimator converges to 0 in
 probability. i.e.
 \begin{eqnarray}
\forall \delta >0~~~\mathbb{P}\left(\frac{1}{n}\|\hat{Y} - X\beta^*
\|_{\ell_2}^2
> \delta \right) \rightarrow 0.
 \end{eqnarray}

\end{definition}

\begin{definition}(Risk consistency or Persistency) Assuming the true model $f^*(X)$ does not have to be
linear, for the regression model with random design,
$(\mathcal{X},\mathcal{Y})\sim F_n \in \mathcal{F}^n$, where
$\mathcal{F}^n$ is a collection of distributions of i.i.d. $m_n+1$
dimensional random vectors. Define the risk function under the
distribution $F_n$ to be $R_{F_n}(\beta) $ (More details in Section
\ref{sec.persistency}). Given a sequence of sets of predictors
$\mathcal{B}_n$, the sequence of estimators
$\hat{\beta}^{\hat{F}_n}\in\mathcal{B}_n$ is called persistent if
for every sequence $F_n \in \mathcal{F}^n$,
\begin{eqnarray}
R_{F_n}(\hat{\beta}^{\hat{F}_n}) - R_{F_n}(\beta_{*}^{F_n})
\stackrel{P}{\rightarrow} 0,
\end{eqnarray}
where
\begin{eqnarray}
\beta_{*}^{F_n} & = & \argmin_{\beta \in \mathcal{B}_n}
R_{F_n}(\beta).
\end{eqnarray}
\end{definition}

For the $\ell_1\text{-}\ell_q$ regularized regression, later, we
will use $\mathcal{B}_n = \{\beta:
\sum_{j=1}^{p_n}(d_j)^{1/q'}\|\beta_j \|^2_{\ell_2} \leq L_n \}$,
for some $\ds L_n = o(\left(n/(\log n)\right)^{1/4})$.

The following table gives a high level summary of our main results,
ordered from very stringent assumptions to much weaker assumptions:
\def\ds{\displaystyle}
\begin{center}
\begin{tabular}{lll}
&&\\
 Variable selection consistency: & $\ds \P\left(S(\hat{\beta}) =
S(\beta^*\right)\rightarrow 1$ &  (R1)\\[15pt]
$\ell_1\text{-}$estimation convergence rate: & $\ds
\|\hat{\beta}-\beta^* \|_{\ell_1} = O_P\left(s_n\bar{d}_n\sqrt{
\frac{\log
m_n}{n}}\right)$ &  (R2)\\[15pt]
Prediction error convergence rate: & $\ds \frac{1}{n}\|\hat{Y} -
X\beta^* \|_{\ell_2}^2 = O_P\left( \frac{s_n\bar{d}_n\log
m_n}{n}\right)$ & (R3)
\\[12pt]
Prediction (misspecified model): & $\ds \frac{1}{n}\| \hat{Y} -
f^*\|_{\ell_2}^2 = O_P\left( \frac{s'\bar{d}_n \log m_n}{n} \right)$
& (R3$^*$)
 \\[12pt]
Persistency (misspecified model):  & $\ds
R_{F_n}(\hat{\beta}^{\hat{F}_n}) - R_{F_n}(\beta_{*}^{F_n})
\stackrel{P}{\rightarrow} 0$
& (R4)\\[15pt]
\end{tabular}
\end{center}
\begin{remark}
(R1) to (R3) assume the true model must be linear, while (R3$^*$)
and (R4) relax this condition so that the model can be misspecified.
Even though (R3) and (R3$^*$) look very similar, (R3$^*$) dropped
the linear model assumption at the price of enforcing another ``{\it
weak sparsity}" condition. Also, (R1), (R2), (R3), and (R3$^*$) are
fixed design results, while (R4) is a random design result.
\end{remark}

In general, the condition for variable selection consistency is the
strongest since it involves not only certain relations among $n$,
$\lambda_n$, $p_n$, $s_n$, $\bar{d}_n$, but also the minimum
absolute value of the parameters, $\rho^*_n=\min_{j\in S}
\|\beta^{*}_j\|_\infty$. The $\ell_1$-estimation consistency and
prediction error consistency requires weaker conditions than
variable selection consistency. Unlike the previous properties, when
the model is misspecified, the prediction error consistency in
(R3$^*$) follows from a sparse oracle inequality. Since both the
sparsity oracle inequalities and persistency does not require the
existence of a true linear model and thus is more general.
Especially, the persistency is about the consistency of the
predictive risk when considering random design and only need a very
weak assumption about the design.

\section{Variable Selection Consistency}\label{sparsistency}

In this Section we study the conditions under which the
$\ell_1\text{-}\ell_q$ estimator is variable selection consistent.
Our proof is adapted from \citep{Wainwright06a_aller} and
\citep{SpAM:07}. The former paper develop the ``witness" proof idea
which is the main framework used in our proof. The latter paper
mainly treat variable selection consistency when $q=2$ in a
nonparametric sparse additive model setting, which makes their
conditions more stringent than ours even when $q=2$.

In the following,  Let $S$ denote the true set of group indices
$\{j: X_j \neq 0 \}$, with $s_n = |S|$, and $S^c$ denote its
complement. Denote $\Lambda_{\min}(C)$ to be the minimum eigenvalue
of the matrix $C$. Then, we have

\begin{theorem}\label{thm.sparsistency} Let $q$ and $q'$ are conjugate exponents with each other, that is $\ds \frac{1}{q} + \frac{1}{q'} = 1$ and $1\leq q,q' \leq \infty$. Suppose that the following conditions hold on the design matrix $X$:
\begin{eqnarray}\label{eqn.sparsistency.condition}
 \Lambda_{\min}\left(\frac{1}{n}X^T_SX_S \right) \geq C_{\min}
> 0 \nonumber  \\
\max_{j\in S^c}\biggl\|(X^T_jX_S) (X^T_SX_S)^{-1}\biggr\|_{q',{q'}}
\leq 1 - \delta, ~~for~some~0 < \delta \leq 1.
\end{eqnarray}
where $\|\cdot\|_{a,b}$ is the matrix norm, defined as $\ds \|A
\|_{a,b} = \sup_{x}\frac{\|Ax\|_{\ell_{b}}}{\|x \|_{\ell_{a}}}$, $1
\leq a,b \leq \infty$. Assume the maximum number of variables with
each group $\bar{d}_n \rightarrow \infty$ and $\bar{d}_n = o(n)$.
Furthermore, suppose the following conditions, which relate the
regularization parameter $\lambda_n$ to the design parameters $n$,
$p_n$, the number of relevant groups $s_n$ and the maximum group
size $\bar{d}_n$:
\begin{eqnarray}
\frac{\lambda_n^2 n}{\log ((p_n-s_n)\bar{d}_n)} \longrightarrow
\infty.\\
\frac{1}{\rho_n^*}\left\{ \sqrt{\frac{\log(s_n\bar{d}_n)}{n}}
+\lambda_n(\bar{d}_n)^{1/q'}\left\|\left(\frac{1}{n}X^T_SX_S
\right)^{-1}\right\|_{\infty,\infty} \right\} \longrightarrow 0.
 \end{eqnarray}
where $\rho_n^* = \min_{j\in S}\|\beta^*_j \|_\infty$. Then, the
$\ell_1\text{-}\ell_q$ regularized regression is variable selection
consistent.
\end{theorem}

\begin{remark}
First, notice that the result established in Theorem
\ref{thm.sparsistency} is a direct generalization of the variable
selection result for Lasso in \citep{Wainwright06a_aller} by setting
$q=1$ and $\bar{d}_n=1$ (as then the $\ell_1\text{-}\ell_q$
degenerates to Lasso). This gives the sufficient conditions for
exact recovery of sparsity pattern in $\beta^{*}$ for the
$\ell_1\text{-}\ell_q$ regularized regression. Also notice that when
$\bar{d}_n$ is bounded from above, the conditions are almost the
same as those of Lasso except the condition in equation
\ref{eqn.sparsistency.condition} which depends on the value of $q$.

Second, we consider the case when $\rho_n$ is bounded away from
zero. Assuming that $q=\infty$ and $\bar{d}_n=n^{1/5}$ (such as in
the fitting of additive model with basis expansion), we must have
$\lambda_n=o(n^{-1/5})$ and as a result of $\ds \frac{\lambda_n^2
n}{\log((p_n-s_n)\bar{d}_n)}\rightarrow \infty$, we need to have
$p_n = o(\exp(n^{3/5}))$. This means that even when we have
increasing group size $\bar{d}_n$, the sparse pattern (in terms of
grouped variables) can still be correctly identified with a large
$p_n$.

Finally, when minimum parameter value $\rho_n\rightarrow 0$, to
ensure variable selection consistency, it can at most converge to
zero at a rate slower than $n^{-1/2}$.
\end{remark}

\noindent{\bf Proof:} Note, the special case when $q=1$ has already
been proved in \citep{Martin:06}. Here, we only consider the case
that $1 < q \leq \infty$.  A vector $\hat{\beta} \in
\mathbb{R}^{m_n}$, $m_n = \sum_{j=1}^{p_n}d_j$,  is an optimum of
the objective function in \eqref{eqn.grouplasso} if and only if
there exists a sequence of  subgradients $\hat{g}_j \in  \partial
\|\hat{\beta}_j \|_{\ell_q}$, such that
\begin{equation}
\label{eq:dbstat} \frac{1}{n}X^T\left(\sum_{j}X_{j}\hat\beta_{j} -
Y\right) + \lambda_{n}(d_j)^{1/q'}\hat{g}_j = \mathbf{0}.
\end{equation}
The subdifferentials $ \partial \|\hat{\beta}_j \|_{\ell_q}$
satisfies the KKT conditions in proposition \ref{proposition.kkt}.

Our argument closely follows the approach of \cite{Martin:06} in the
linear case. In particular, we proceed by a ``witness'' proof
technique, to show the existence of a coefficient-subgradient pair
$(\hat{\beta}, \hat{g})$ for which $\supp(\hat{\beta}) =
\supp(\beta^*)$. To do so, we first set $\hat{\beta}_{\Sc} = \bm{0}$
and $\hat{g}_{S}$ to be the vector concatenating all the subvectors
$\hat{g}_j$'s, for $j \in S$. We also define $\hat{g}_{S^c}$ and
$\hat{\beta}_S$ in a similar way. And we then obtain
$\hat{\beta}_{S}$ and $\hat{g}_{\Sc}$ from the stationary conditions
in~\eqref{eq:dbstat}. By showing that, with high probability,
\begin{eqnarray}
\hat{\beta}_{j} &\neq& \mathbf{0} \;\; \text{for $j \in S$}\\[5pt]
\hat{g}_j &\in& B^{q'}(1) \;\;\text{for $j \in \Sc$},
\end{eqnarray}
this demonstrates that with high probability there exists an optimal
solution to the optimization problem in~\eqref{eqn.grouplasso} that
has the same sparsity pattern as the true model.

Setting $\hat{\beta}_{\Sc} = \mathbf{0}$ and
\begin{eqnarray}
\hat{g}_{j} = \left\{ \begin{array}{cc}
                        \ds \Biggl\{\biggl(
\frac{|{\hat{\beta_j}}_{\ell}|^{q-1}\mathrm{sign}({\hat{\beta_j}}_{\ell})}{\|\hat{\beta_j}\|^{{q-1}}_{\ell_q}}
\biggr)_{\ell=1}^{d_j} \Biggr\} & 1 < q < \infty \\
                        \ds \mathrm{conv}\{\mathrm{sign}({\hat{\beta_j}}_\ell)e_\ell:|{\hat{\beta_j}}_\ell| = \|\hat{\beta_j} \|_{\ell_\infty} \} &
                        q =\infty
                      \end{array}
\right.
\end{eqnarray}
for $j\in S$, denote $W= \mathrm{diag}((d_1)^{1/q'}I_{d_1},\ldots,
(d_{p})^{1/q'}I_{d_p})$ where $I_{d_j}$ is a $d_j \times d_j$
identity matrix. We define $W_S$ to be submatrix of $W$ by
extracting out the rows and columns corresponding to the group index
set $S$. The stationary condition for $\hat{\beta}_{S}$ is
\begin{equation}
\frac{1}{n}X_{S}^T\left(X_{S}\hat{\beta}_{S} - Y\right) +
\lambda_{n}W_S\hat g_S = \mathbf{0}.
\end{equation}
Let $\epsilon = (\epsilon_1,\ldots, \epsilon_n)^T$,  then the
stationary condition can be written as
\begin{eqnarray}
\frac{1}{n}X_S^T X_S \left(\hat{\beta}_S - \beta_S^*\right)  -
 \frac{1}{n}X_S^T \epsilon  + \lambda_nW_S\hat g_S = \mathbf{0}
\end{eqnarray}
or
\begin{eqnarray}
\label{eq:bseq} \hat{\beta}_S - \beta_S^* = \left( \frac{1}{n}X_S^T
X_S\right)^{-1} \left(\frac{1}{n}X_S^T \epsilon  - \lambda_nW_S\hat
g_S\right)
\end{eqnarray}
assuming that $\ds \frac{1}{n}X_S^T X_S$ is nonsingular. Recalling
our definition
\begin{equation}
\rho_n^* = \min_{j \in S}\|\beta_j^*\|_{\ell_\infty} > 0.
\end{equation}
it suffices to show that
\begin{equation}
\|\hat{\beta}_{S} - \beta^{*}_{S}\|_{\ell_\infty} <
\frac{\rho_n^*}{2}
\end{equation}
in order to ensure that $\supp(\beta_{S}^*) = \supp(\hat{\beta}_S) =
\left\{j\,:\, \|\hat{\beta}_j\|_{\ell_\infty} \neq 0\right\}$.

Using $\ds \Sigma_{SS} = \frac{1}{n}X_{S}^TX_{S}$ to simplify
notation, we have the $\ell_\infty$ bound
\begin{eqnarray}
\label{eq:linftybnd} \|\hat{\beta}_{S} -
\beta^{*}_{S}\|_{\ell_\infty} \leq
\left\|\Sigma_{SS}^{-1}\left(\frac{1}{n}
X_{S}^T\epsilon\right)\right\|_{\ell_\infty} + \lambda_n
\left\|\Sigma_{SS}^{-1}W_S\hat g_{S}\right\|_{\ell_\infty}.
\end{eqnarray}
We now proceed to bound the quantities above.  First note that for
$j\in S$, $\|\hat{g}_j\|_{\ell_{q'}} = 1$. Therefore, since
\begin{equation}
\|\hat{g}_{S}\|_{\ell_\infty} = \max_{j \in
S}\|\hat{g}_{j}\|_{\ell_\infty} \le \max_{j \in
S}\|\hat{g}_{j}\|_{\ell_{q'}} = 1
\end{equation}
we have that
\begin{equation}
\left\|\Sigma_{SS}^{-1}W_S \hat g_S\right\|_{\ell_\infty} \le
(\bar{d}_n)^{1/q'} \left\|\Sigma_{SS}^{-1}\right\|_{\infty,\infty} .
\end{equation}
Therefore
\begin{eqnarray}
\nonumber \|\hat{\beta}_{S} - \beta^{*}_{S}\|_{\ell_\infty} \leq
\left\|\Sigma_{SS}^{-1}\left(\frac{1}{n}X_{S}^T\epsilon\right)\right\|_{\ell_\infty}
+
\lambda_n(\bar{d}_n)^{1/q'}\left\|\Sigma_{SS}^{-1}\right\|_{\infty,\infty}.
\label{eq:linftybnd2}
\end{eqnarray}

Finally, consider $Z = \ds
\Sigma_{SS}^{-1}\left(\frac{1}{n}X_{S}^T\epsilon\right)$. Note that
$\epsilon \sim N(0,\sigma^{2}I)$, so that $Z$ is Gaussian as well,
with mean zero. Consider its $\ell$-th component,
$Z_{\underline{\ell}} = e_{\ell}^TZ$. Then $\E[Z_{\underline{\ell}}]
= 0$, and
\begin{eqnarray}
\Var(Z_{\underline{\ell}}) &=&
\frac{\sigma^{2}}{n}e_{\ell}^T\Sigma_{SS}^{-1}e_{\ell} \le
\frac{\sigma^{2}}{nC_{\min}}.
\end{eqnarray}
By the comparison results on Gaussian maxima
\citep{Ledoux:Talagrand:91}, we have then that
\begin{equation}
\E\left[\|Z\|_{\ell_\infty}\right] \;\le\; 3 \sqrt{\log(s\bar{d}_n)
}\max_{\underline{\ell}}\sqrt{\mathrm{Var}(Z_{\underline{\ell}})}
\;\le\; 3\sigma \sqrt{\frac{\log(s\bar{d}_n)}{nC_{\min}}}.
\end{equation}
An application of Markov's inequality then gives that
\begin{eqnarray}
\nonumber \P\left(\| \hat{\beta}_{S} - \beta^{*}_{S}\|_{\ell_\infty}
>
    \frac{\rho_n^*}{2}\right) & \le &
 \P\left(\|Z\|_{\ell_\infty} + \lambda_n(\bar{d}_n)^{1/q'}\left\|\Sigma_{SS}^{-1}\right\|_{\infty,\infty} > \frac{\rho_n^*}{2} \right) \\
&\le& \frac{2}{\rho_n^*}\left\{\E\left[\|Z\|_{\ell_\infty}\right] +
\lambda_n(\bar{d}_n)^{1/q'}\left\|\Sigma_{SS}^{-1}\right\|_{\infty,\infty} \right\} \\
&\le& \frac{2}{\rho_n^*}\left\{ 3 \sigma
\sqrt{\frac{\log(s\bar{d}_n)}{n C_{\min}}} +
\lambda_n(\bar{d}_n)^{1/q'}\left\|\Sigma_{SS}^{-1}\right\|_{\infty,\infty}\right\}
\end{eqnarray}
which converges to zero under the condition that
\begin{equation}
\label{eq:thiscond} \frac{1}{\rho_n^*}\left\{
\sqrt{\frac{\log(s\bar{d}_n)}{n}}
+\lambda_n(\bar{d}_n)^{1/q'}\left\|\Sigma_{SS}^{-1}\right\|_{\infty,\infty}
\right\} \longrightarrow 0.
\end{equation}

We now analyze $\hat{g}_{\Sc}$. Recall that we have set $\hat
\beta_{S^c} = \beta^*_{S^c} = \bm{0}$.  The stationary condition for
$j \in \Sc$ is thus given by
\begin{equation}
\frac{1}{n}X_{j}^T\left(X_{S}\hat \beta_{S} -
  X_{S}\beta^{*}_{S} - \epsilon \right) + \lambda_{n}(d_j)^{1/q'}\hat g_{j} = \mathbf{0}.
\end{equation}
Therefore,
\begin{eqnarray}
\nonumber \hat g_{S^c} &=& \frac{W^{-1}_{S^c}}{\lambda_n} \left\{
\frac{1}{n}X_{S^c}^TX_{S} \left(\beta^*_{S} - \hat\beta_{S}\right)
    + \frac{1}{n}X_{S^c}^T\epsilon \right\} \\
&=& \frac{W^{-1}_{S^c}}{\lambda_n} \left\{
\frac{1}{n}X_{S^c}^TX_{S}\left( \frac{1}{n}X_S^T X_S\right)^{-1}
\left(\lambda_nW_S\hat g_S-\frac{1}{n}X_S^T \epsilon \right)
    +  \frac{1}{n}X_{S^c}^T \epsilon \right\}
\nonumber
\\
&=& \frac{W^{-1}_{S^c}}{\lambda_n} \left\{ \Sigma_{S^c S} \Sigma_{S
S}^{-1} \left(\lambda_n W_S\hat g_S - \frac{1}{n} X_S^T \epsilon
\right)
    + \frac{1}{n}X_{S^c}^T\epsilon \right\}
\label{eq:g}
\end{eqnarray}
from equation \eqref{eq:bseq}.

We want to show that
\begin{equation}
\hat{g}_j \in B^{q'}(1)
\end{equation}
for all $j\in S^c$. From \eqref{eq:g}, we see that $\hat g_j$ is
Gaussian, with mean
\begin{eqnarray}
\mu_j = \E(\hat g_j) =
(d_j)^{-1/q'}\Sigma_{jS}\Sigma_{SS}^{-1}W_S\hat g_S .
\end{eqnarray}
We then obtain the bound
\begin{eqnarray}
\nonumber \|\mu_j\|_{\ell_{q'}} &\leq&
\left\|\Sigma_{jS}\Sigma_{SS}^{-1}\right\|_{q',{q'}} \|
\hat{g}_S\|_{\ell_{q'}} =
\left\|\Sigma_{jS}\Sigma_{SS}^{-1}\right\|_{q',{q'}} \leq
1-\delta~~~ \mathrm{~for~ some~} \delta > 0.
\end{eqnarray}

It therefore suffices to show that
\begin{eqnarray}
\label{eq:toshow} \P\left( \max_{j\in S^c} {({d}_j)^{1/q'}}\| \hat
g_j - \mu_j\|_{\ell_\infty} > \frac{\delta}{2} \right)
\longrightarrow 0
\end{eqnarray}
since this implies that
\begin{eqnarray}
\|\hat g_j \|_{\ell_{q'}} &\leq&\|\mu_j\|_{\ell_{q'}} + \|\hat g_j - \mu_j\|_{\ell_{q'}} \\
  &\leq& \|\mu_j\|_{\ell_{q'}} + {({d}_j)^{1/q'}} \|\hat g_j - \mu_j\|_{\ell_\infty} \\
  &\leq&  (1-\delta) + \frac{\delta}{2} + o(1)
\end{eqnarray}
with probability approaching one.  
To show \eqref{eq:toshow}, we again appeal to comparison results of
Gaussian maxima. Define
\begin{eqnarray}
Z_j = ({d}_j)^{1/q'}\lambda_n(\hat{g}_j - \mu_j) \;=\; X_j^T\left(I
- X_S (X_S^T X_S)^{-1} X_S^T\right) \frac{\epsilon}{n}
\end{eqnarray}
for $j\in S^c$.  Then $Z_j$ are zero mean Gaussian random vector,
and we need to show that
\begin{eqnarray}
\P\left(\max_{j\in S^c} \frac{\|Z_j\|_{\ell_\infty}}{\lambda_n} \geq
\frac{\delta}{2}\right) \longrightarrow \infty.
\end{eqnarray}
Let $Z_{jk}$ represent the $k$-th element of $Z_j$ for $j \in S^c$. A calculation shows that $\E(Z_{jk}^2) \leq \ds \frac{\sigma^2}{n}$.  
 Therefore, we have
by Markov's inequality and the  comparison results of Gaussian
maxima that
\begin{eqnarray}
\nonumber \P\left(\max_{j\in S^c}
\frac{\|Z_j\|_{\ell_\infty}}{\lambda_n} \geq \frac{\delta}{2}\right)
& \leq &\frac{2}{\delta\lambda_n} \, \E\left(\max_{{j\in S^c},k} |Z_{jk}|\right) \\
&\leq& \frac{2}{\delta\lambda_n} \left( 3
\sqrt{\log((p_n-s_n)\bar{d}_n)} \max_{{j\in S^c},k}
  \sqrt{\E\left( Z_{jk}^2\right)}\right) \\
&\leq& \frac{6\sigma}{\delta\lambda_n} \sqrt{\frac{\log((p_n-s_n)
\bar{d}_n)}{n}}
\end{eqnarray}
which converges to zero under the condition that
\begin{eqnarray}
\frac{\lambda_n^2 n}{ \log ((p_n -s_n)\bar{d}_n)} \longrightarrow
\infty.
\end{eqnarray}
This is just the condition  in the statement of the theorem. $\Box$

\section{Estimation Consistency}\label{sec.consistency}

In this section, we  prove the estimation consistency results under
two types assumptions: (i) When the model is correctly specified,
i.e., the true model is linear, we can achieve both
$\ell_1\text{-}$consistency results and derive the optimal rate of
convergence for the prediction error. (ii) When the model is
misspecified, i.e. the true model is not linear, we can still
achieve a sparsity oracle inequality, which provide a bound of the
prediction error using the loss of the prediction oracle with the
number of nonzero groups of the prediction loss involved in. Under
the ``{\it weak sparsity}" condition, we can still obtain a rate of
convergence of the prediction error which is similar to the
convergence rate obtained under the linear model assumption.

We begin with a technical lemma, which is essentially lemma 1 as in
\citep{Bunea:annal:07} and \citep{Bick:Ritov:2007}, but need to be
extended to handle the group structures in the more general
$\ell_1\text{-}\ell_q$ regularized regression setting.

\begin{lemma}\label{lemma.inequality}
Let $\epsilon_1\ldots,\epsilon_n$ be independent
$\mathcal{N}(0,\sigma^2)$ random variables with $\sigma^2>0$ and Let
$\hat{Y}=X\hat{\beta}$ be the $\ell_1\text{-}\ell_q$ regularized
regression estimator with $1\leq q \leq \infty$ as in
(\ref{eqn.grouplasso}) with
\begin{eqnarray}
\lambda_n = A\sigma\sqrt{\frac{\log m_n}{n}}
\end{eqnarray}
for some $A > 2\sqrt{2}$. Then, for all $m_n \geq 2$, $n>1$, with
probability of at least $1 - {m_n}^{1-A^2/8}$ we have simultaneously
for all $\beta \in \mathbb{R}^{m_n}$:
\begin{eqnarray}\label{eq.inequality}
\frac{1}{n}\|\hat{Y} - X\beta^* \|_{\ell_2}^2 +
\lambda_n\sum_{j=1}^{p_n}(d_j)^{1/q'}\|\hat{\beta}_j-\beta_j\|_{\ell_q}\leq
\frac{1}{n}\|X\beta - X\beta^*\|^2_{\ell_2} + 4\sum_{j\in
S(\beta)}\lambda_n{(d_j)^{ 1/q'}}\|\hat{\beta}_j - \beta_j
\|_{\ell_q}
\end{eqnarray}
where $S(\beta)$ denotes the set of nonzero group indices of
$\beta$.
\end{lemma}

\noindent{\bf Proof:} By the definition of $\hat{Y} = X\hat{\beta}$,
we have
\begin{eqnarray}
\frac{1}{2n}\|Y - X\hat{\beta} \|_{\ell_2}^2 + \lambda_n
\sum_{j=1}^{p_n}{(d_j)}^{1/q'}\|\hat{\beta}_j\|_{\ell_{q}} \leq
\frac{1}{2n}\|Y - X{\beta} \|_{\ell_2}^2+ \lambda_n
\sum_{j=1}^{p_n}{(d_j)^{1/q'}}\|\beta_j\|_{\ell_{q}} \nonumber
\end{eqnarray}
for all $\beta \in \mathbb{R}^{m_n}$, $m_n = \sum_{j=1}^{p_n}d_j$,
which we may rewritten as
\begin{eqnarray}
\lefteqn{\frac{1}{n}\|X\beta^* - X\hat{\beta} \|_{\ell_2}^2 +
2\lambda_n \sum_{j=1}^{p_n}{(d_j)^{1/q'}}\|\hat{\beta}_j\|_{\ell_q}}
&& \nonumber \\ && \leq \frac{1}{n}\|X\beta^* - X{\beta}
\|_{\ell_2}^2 + 2\lambda_n
\sum_{j=1}^{p_n}{(d_j)^{1/q'}}\|\beta_j\|_{\ell_q} +
\frac{2}{n}\epsilon^TX(\hat{\beta}-\beta).
\end{eqnarray}
For each $\underline{j} = 1,\ldots, m_n$, we define the random
variables $\ds V_{\underline{j}} =
\frac{1}{n}X^T_{\underline{j}}\epsilon$, and the event
$$\ds \mathcal{A} =
\bigcap_{\underline{j}=1}^{m_n}\left\{2|V_{\underline{j}}| \leq
\lambda_n\right\}. $$ Under the normality assumption, we have that
\begin{eqnarray}
\sqrt{n}V_{\underline{j}} \sim
\mathcal{N}(0,\sigma^2)~~~\underline{j} = 1,\ldots, m_n.
\end{eqnarray}
Using the elementary bound on the tails of Gaussian distribution we
find that the probability of the complementary event $\mathcal{A}^c$
satisfies
\begin{eqnarray}
\mathbb{P}\{\mathcal{A}^c \}  & \leq &
\sum_{\underline{j}=1}^{m_n}\mathbb{P}\{\sqrt{n}|V_{\underline{j}}|
> \sqrt{n}\lambda_n/2\}
 \leq  m_n\mathbb{P}\{|Z| \geq
\sqrt{n}\lambda_n/(2\sigma)) \} \\
& \leq & m_n\exp\left(-\frac{n\lambda_n^2}{8\sigma^2} \right) =
m_n\exp\left(-\frac{A^2\log m_n}{8} \right) = m_n^{1-A^2/8}
\end{eqnarray}
where $Z \sim \mathcal{N}(0,1)$. Then, on the set $\mathcal{A}$, we
have
\begin{eqnarray}
 \frac{2}{n}\epsilon^TX(\hat{\beta}-\beta)& = & 2\sum_{\underline{j}=1}^{m_n}V_{\underline{j}}(\hat{\beta}_{\underline{j}} - \beta_{\underline{j}})
\leq \sum_{\underline{j}=1}^{m_n}
\lambda_n|\hat{\beta}_{\underline{j}} - \beta_{\underline{j}}|\leq
\sum_{j=1}^{p_n} \lambda_n{(d_j)^{1/q'}}\|\hat{\beta}_j -
\beta_j\|_{\ell_q} \nonumber
\end{eqnarray}
and therefore, still on the set $\mathcal{A}$,
\begin{eqnarray}
\lefteqn{\frac{1}{n}\|X\beta^* - X\hat{\beta} \|_{\ell_2}^2 \leq
\frac{1}{n}\|X\beta^* - X{\beta} \|_{\ell_2}^2} && \nonumber \\ && +
2\lambda_n \sum_{j=1}^{p_n}{(d_j)^{1/q'}}\|\beta_j\|_{\ell_q} +
\sum_{j=1}^{p_n} \lambda_n{(d_j)^{1/q'}}\|\hat{\beta}_j -
\beta_j\|_{\ell_q}- 2\lambda_n
\sum_{j=1}^{p_n}{(d_j)^{1/q'}}\|\hat{\beta}_j\|_{\ell_q}.
\end{eqnarray}
Adding the same term $\ds \sum_{j=1}^{p_n}
\lambda_n{(d_j)^{1/q'}}\|\hat{\beta}_j - \beta_j\|_{\ell_q}$ on both
sides, we obtain
\begin{eqnarray}
\lefteqn{\frac{1}{n}\|X\beta^* - X\hat{\beta} \|_{\ell_2}^2 +
\lambda_n\sum_{j=1}^{p_n} {(d_j)^{1/q'}}\|\hat{\beta}_j -
\beta_j\|_{\ell_q} \leq \frac{1}{n}\|X\beta^* - X{\beta}
\|_{\ell_2}^2} && \nonumber \\ && + 2\lambda_n
\sum_{j=1}^{p_n}{(d_j)^{1/q'}}\|\beta_j\|_{\ell_q} +
2\lambda_n\sum_{j=1}^{p_n} {(d_j)^{1/q'}}\|\hat{\beta}_j -
\beta_j\|_{\ell_q}- 2\lambda_n
\sum_{j=1}^{p_n}{(d_j)^{1/q'}}\|\hat{\beta}_j\|_{\ell_q}.
\end{eqnarray}

Recall $S(\beta)$ to be the set of non-zero group indices of
$\beta$. Rewriting the right-hand side of the previous display,
then, on set $\mathcal{A}$
\begin{eqnarray}
\lefteqn{\frac{1}{n}\|X\beta^* - X\hat{\beta} \|_{\ell_2}^2 +
\lambda_n\sum_{j=1}^{p_n} {(d_j)^{1/q'}}\|\hat{\beta}_j -
\beta_j\|_{\ell_q}} && \nonumber \\ && \leq \frac{1}{n}\|X\beta^* -
X{\beta} \|_{\ell_2}^2 + 2\left( \sum_{j=1}^{p_n} \lambda_n(d_j)^{
1/q'}\|\hat{\beta}_j - \beta_j \|_{\ell_q} - \sum_{j\notin
S(\beta)}\lambda_n{(d_j)^{1/q'}}\|\hat{\beta}_j\|_{\ell_q}\right) \nonumber\\
\nonumber & ~ &+2\left(  \sum_{j\in
S(\beta)}\lambda_n(d_j)^{1/q'}\|{\beta}_j\|_{\ell_q} - \sum_{j\in
S(\beta)}\lambda_n{(d_j)^{1/q'}}\|\hat{\beta}_j\|_{\ell_q}\right) \\
\nonumber & \leq &  \frac{1}{n}\|X\beta - X\beta^*\|^2_{\ell_2} +
4\sum_{j\in S(\beta)}\lambda_n{(d_j)^{ 1/q'}}\|\hat{\beta}_j -
\beta_j \|_{\ell_q}
\end{eqnarray}
by the triangle inequality and the fact that $\beta_j = \mathbf{0}$
for $j \notin S(\beta)$. $\Box$

\subsection{Estimation Consistency Under the Linear Model Assumption}

Assuming the true model is linear, to obtain the
$\ell_1\text{-}$consistency result, a key assumption on the design
matrix is needed, which is stated as the following
\begin{assumption}\label{assumption.re1}
Recall that $s_n = S(\beta^*)$, assume for any vector $\gamma \in
\mathbb{R}^{m_n}$ satisfies
\begin{eqnarray}
\ds \kappa \equiv \ds \min_{S_0 \subset \{1,\ldots, p \}: |S_0|\leq
s_n }~~ \min_{ \sum_{j \in S_0^c}(d_j)^{1/q'}\|\gamma_j\|_{\ell_q}
\leq 3\sum_{j \in S_0}(d_j)^{1/q'}\|\gamma_j\|_{\ell_q}
}\frac{\|X\gamma \|_{\ell_2}}{\sqrt{n}\sqrt{\sum_{j\in
S_0}(d_j)^{2/q'-1}\|\gamma_j \|^2_{\ell_q}}}
> 0.
\end{eqnarray}
\end{assumption}
\begin{remark}
Before proving the following theorem, we pause to make some comments
about this assumption.

First, For $q=1$ (thus, $q'=\infty$), this assumption is very
similar to the {\it restricted eigenvalue} assumption as in
\citep{Bick:Ritov:2007}, which is defined as
\begin{eqnarray}
\ds \kappa \equiv \ds \min_{S_0 \subset \{1,\ldots, p \}: |S_0|\leq
s_n }~~ \min_{ \sum_{j \in S_0^c}\|\gamma_j\|_{\ell_1} \leq 3\sum_{j
\in S_0}\|\gamma_j\|_{\ell_1} }\frac{\|X\gamma
\|_{\ell_2}}{\sqrt{n}\sqrt{\sum_{j\in S_0}\|\gamma_j \|^2_{\ell_2}}}
> 0.
\end{eqnarray}
However, our assumption is slightly weaker, due to the fact that,
for any $\gamma \in \mathbb{R}^{d_j}$
\begin{eqnarray}
\|\gamma_j \|^2_{\ell_1} \leq d_j\|\gamma_j \|^2_{\ell_2}.
\end{eqnarray}

Second, the quantity $\sqrt{\sum_{j\in S_0}(d_j)^{2/q'-1}\|\gamma_j
\|^2_{\ell_q}}$ in our assumption balances between $q=1$ and
$q=\infty$. For example, when $q=1$, $\|\gamma_j \|^2_{\ell_1}$ is
relatively large, but $(d_j)^{2/q'-1} = (d_j)^{-1}$  is very small.
While for $q=\infty$, $\|\gamma_j \|^2_{\ell_q} =  \|\gamma_j
\|^2_{\ell_\infty}$ is relatively small, however, $(d_j)^{2/q'-1} =
(d_j)^{1}$ is very significant. In this sense, $q=2$ seems the most
balanced one, due to the fact that
\begin{eqnarray}
\sum_{j\in S_0}(d_j)^{-1}\|\gamma_j \|^2_{\ell_1} \leq \sum_{j\in
S_0}\sqrt{d_j}\|\gamma_j \|^2_{\ell_2} \leq \sum_{j\in
S_0}d_j\|\gamma_j \|^2_{\ell_\infty}
\end{eqnarray}
Therefore, among $q=1,2,\infty$, $q =2 $ needs the weakest
assumption, this provides some insights about why group Lasso might
also be a suitable choice for grouped variable selection. However,
we need to more cautions to say which value of $q$ is the best.
Since in real applications, the choice of $q$ might depends on the
true relevant coefficients $\beta^*_S$. If different components in
the relevant groups are on the same order of magnitude, $q=\infty$
might be more suitable, on the contrary, if some relevant
coefficients are very small relative to the others, $q=1$ might be
better. we plan to investigate this issue in  a separate paper.

\end{remark}
\begin{theorem}\label{theorem.rate} (Estimation consistency under linear model assumptions) Under assumption \ref{assumption.re}, let $\epsilon_1, \ldots,
\epsilon_n$ be independent $\mathcal{N}(0,\sigma^2)$ random
variables with $\sigma^2 > 0$. Consider the $\ell_1\text{-}\ell_q$
regularized estimator defined by (\ref{eqn.grouplasso}) with
\begin{eqnarray}
\lambda_n = A\sigma\sqrt{\frac{\log m_n}{n}}
\end{eqnarray}
for some $A > 2\sqrt{2}$. then, for all $n \geq 1$ with probability
at least $\ds 1 - {m_n}^{1-A^2/8}$ we have
\begin{eqnarray}
\frac{1}{n}\|\hat{Y} - X\beta^* \|_{\ell_2}^2 \leq
\frac{9A^2\sigma^2}{\kappa^2}  \frac{s_n\bar{d}_n\log m_n}{n}\\
\|\hat{\beta} - \beta^* \|_{\ell_1} \leq
\frac{12A^2\sigma^2s_n\bar{d}_n}{\kappa^2}\sqrt{\frac{\log m_n}{n}}.
\end{eqnarray}
\end{theorem}

\begin{remark}
From this theorem, we obtain $\ell_1\text{-}$consistency and the
corresponding rate of convergence. Due to the fact that $\|
\gamma\|_{\ell_q} \leq \|\gamma\|_{\ell_1}$ for all $1< q\leq
\infty$, we obtain $\ell_q$ consistency also if $\ds\
s_n\bar{d}_n\sqrt{\frac{\log m_n}{n}}\rightarrow 0.$  If we want to
the rate of convergence for $\ell_2\text{-}$consistency, a direct
result will be
\begin{eqnarray}
\|\hat{\beta} - \beta^* \|^2_{\ell_2} \leq
\frac{144A^4\sigma^4s^2_n\bar{d}^2_n}{\kappa^4}\frac{\log m_n}{n}.
\end{eqnarray}
which is suboptimal. Recall that $\|\hat{\beta} - \beta^*
\|^2_{\ell_1} \leq p_n\bar{d}_n\|\hat{\beta} - \beta^*
\|^2_{\ell_2}$, if $|S(\hat{\beta})|$ is $O(s_n)$ and the elements
in  $\hat{\beta}_j-\beta^*_j$ are balanced for $j \in S$, then we
can also achieve the optimal rate of convergence for
$\ell_2\text{-}$norm consistency. How to obtain optimal rate of
convergence for $\ell_q\text{-}$consistency for general $q$ would be
an interesting future work.
\end{remark}

\noindent{\bf Proof:} From equation \ref{eq.inequality}, Using
$\beta = \beta^*$, we have that on the event $\mathcal{A}$,
\begin{eqnarray} \label{eq.1}
\frac{1}{n}\|\hat{Y} - X\beta^* \|_{\ell_2}^2 \leq 3\sum_{j\in
S(\beta^*)}\lambda_n{(d_j)^{ 1/q'}}\|\hat{\beta}_j - \beta^*_j
\|_{\ell_q} \leq 3\lambda_n\sqrt{\bar{d}_ns_n} \sqrt{\sum_{j \in
S(\beta^*)
}(d_j)^{2/q'-1}\|\hat{\beta}_j - \beta^*_j \|^2_{\ell_q}} \label{eq.predictiverate}\\
\sum_{j\in S(\beta^*)^c}(d_j)^{1/q'}\|\hat{\beta}_j - \beta^*_j
\|_{\ell_q} \leq 3\sum_{j\in S(\beta^*)}{(d_j)^{
1/q'}}\|\hat{\beta}_j - \beta^*_j \|_{\ell_q}. \label{eq.L1rate}
\end{eqnarray}
By the last equation, we have that assumption \ref{assumption.re1}
hold on event $\mathcal{A}$, by this assumption, we have that
\begin{eqnarray}\label{eq.2}
\frac{1}{n}\|\hat{Y} - X\beta^* \|_{\ell_2}^2 \geq\kappa^2\sum_{j
\in S(\beta^*) }(d_j)^{2/q'-1}\|\hat{\beta}_j - \beta^*_j
\|^2_{\ell_q}.
\end{eqnarray}
By combining the above inequalities, we get
\begin{eqnarray}
\frac{1}{n}\|\hat{Y} - X\beta^* \|_{\ell_2}^2 \leq
\frac{9\lambda^2_ns_n\bar{d}_n}{\kappa^2} \label{eq.rate}
\end{eqnarray}
and
\begin{eqnarray}
 \sqrt{\sum_{j \in
S(\beta^*) }(d_j)^{2/q'-1}\|\hat{\beta}_j - \beta^*_j \|^2_{\ell_q}}
\leq \frac{3\lambda_n\sqrt{\bar{d}_ns_n}}{\kappa^2}.
\end{eqnarray}
Thus, we have
\begin{eqnarray}
\|\hat{\beta} - \beta^* \|_{\ell_1} & = & \sum_{j=1}^{p_n}\|
\hat{\beta} - \beta^*\|_{\ell_1}  \leq
\sum_{j=1}^{p_n}(d_j)^{1/q'}\|\hat{\beta}_j-\beta^*_j \|_{\ell_q} \\
& = & \sum_{j \in S(\beta^*)}(d_j)^{1/q'}\|\hat{\beta}_j-\beta^*_j
\|_{\ell_q} + \sum_{j\in
S(\beta^*)^c}(d_j)^{1/q'}\|\hat{\beta}_j-\beta^*_j \|_{\ell_q} \\
& \leq & 4\sum_{j \in
S(\beta^*)}(d_j)^{1/q'}\|\hat{\beta}_j-\beta^*_j \|_{\ell_q}
 \leq   4\sqrt{\bar{d}_ns_n} \sqrt{\sum_{j \in S(\beta^*)
}(d_j)^{2/q'-1}\|\hat{\beta}_j - \beta^*_j \|^2_{\ell_q}} \\
& \leq & \frac{12\lambda_n\bar{d}_ns_n}{\kappa^2} =
\frac{12A^2\sigma^2s_n\bar{d}_n}{\kappa^2}\sqrt{\frac{\log m_n}{n}}.
\end{eqnarray}
Note, equation \ref{eq.rate} is exactly equation
\ref{eq.predictiverate}. $\Box$

\subsection{Oracle Inequalities for Prediction Error Under Misspecified Models}

Assuming the true regression function $f^*(X)$ is not linear, i.e.
the model is misspecified. We can no longer obtain the optimal rate
of convergence directly. But we can still obtain a sparsity oracle
inequality, which can bound the prediction error in terms of nonzero
components of the prediction oracle.

\begin{assumption}\label{assumption.re} Assume $s'$
 is an integer such that $1 \leq s' \leq p_n$, and $\delta$ is some positive number,
 then, for any $\gamma \neq 0$
 \begin{eqnarray}
\ds \kappa(s',\delta) \equiv \ds \min_{S_0 \subset \{1,\ldots, p \}:
|S_0|\leq s' }~\min_{ \sum_{j \in
S_0^c}(d_j)^{1/q'}\|\gamma_j\|_{\ell_q} \leq (2 +
\frac{3}{\delta})\sum_{j \in S_0}(d_j)^{1/q'}\|\gamma_j\|_{\ell_q}
}\frac{\|X\gamma \|_{\ell_2}}{\sqrt{n}\sqrt{\sum_{j\in
S_0}(d_j)^{2/q'-1}\|\gamma_j \|^2_{\ell_q}}}
> 0. \nonumber
\end{eqnarray}
\end{assumption}

\begin{theorem}\label{thm.oracle}
Under assumption (\ref{assumption.re}), let $\epsilon_1, \ldots,
\epsilon_n$ be independent $\mathcal{N}(0,\sigma^2)$ random
variables with $\sigma^2 > 0$. Consider the $\ell_1\text{-}\ell_q$
regularized estimator defined by (\ref{eqn.grouplasso}) with
\begin{eqnarray}
\lambda_n = A\sigma\sqrt{\frac{\log m_n}{n}}
\end{eqnarray}
for some $A > 2\sqrt{2}$. then, for all $n \geq 1$ with probability
at least $\ds 1 - {m_n}^{1-A^2/8}$ we have
\begin{eqnarray}
\lefteqn{\frac{1}{n}\|f^* - X\hat{\beta} \|_{\ell_2}^2 } &&
\nonumber
\\ && \leq (1 + \delta)\inf_{\beta \in \mathbb{R}^{m_n}:
|S(\beta)|\leq s'}\biggl\{\frac{1}{n}\|f^* - X{\beta} \|_{\ell_2}^2
+ \frac{C(\delta)A^2\sigma^2}{\kappa(s',\delta)^2}\left(
\frac{\bar{d}_n|S(\beta)|\log m_n}{n}\right)\biggr\}
\end{eqnarray}
where $C(\delta) > 0$ is a constant depending only on $\delta$.
While $|S(\beta)|$ represents the number of nonzero elements in the
set $S(\beta)$.
\end{theorem}
\begin{remark}
From this sparsity oracle inequality, if we add some assumptions,
such as  there exists some $\beta'$, such that$\ds \frac{1}{n}\|f^*
- X{\beta'} \|_{\ell_2}^2 \rightarrow 0$, then we can still obtain
prediction error consistency if $\ds \frac{\bar{d}_n|S(\beta')|\log
m_n}{n} \rightarrow 0$. If we also want to obtain a convergence rate
similar to that as in theorem \ref{theorem.rate}, more conditions
will be needed, as is shown in corollary \ref{corollary.rate1}.
\end{remark}

\noindent{\bf Proof:} Fix an arbitrary $\beta \in \mathbb{R}^{m_n}$
with $|S(\beta)|\leq s'$.  On the event $\mathcal{A}$, we get from
lemma ~\ref{lemma.inequality} that
\begin{eqnarray}\label{eq:sparseoracle}
\lefteqn{\frac{1}{n}\|\hat{Y} - f^* \|_{\ell_2}^2 +
\lambda_n\sum_{j=1}^{p_n}(d_j)^{1/q'}\|\hat{\beta}_j-\beta_j\|_{\ell_q}}
&& \nonumber \\ && \leq \frac{1}{n}\|X\beta - f^*\|^2_{\ell_2} +
4\sum_{j\in S(\beta)}\lambda_n{(d_j)^{ 1/q'}}\|\hat{\beta}_j -
\beta_j \|_{\ell_q}
\end{eqnarray}
Further from above, we can get that
\begin{eqnarray}
\frac{1}{n}\|\hat{Y} - f^* \|_{\ell_2}^2  & \leq &
\frac{1}{n}\|X\beta - f^*\|^2_{\ell_2} + 3\lambda_n\sum_{j\in
S(\beta)}{(d_j)^{
1/q'}}\|\hat{\beta}_j - \beta_j \|_{\ell_q} \\
 & \leq &
\frac{1}{n}\|X\beta - f^*\|^2_{\ell_2} +
3\lambda_n\sqrt{\bar{d}_n|S(\beta)|}\sqrt{\sum_{j \in S(\beta)
}(d_j)^{2/q'-1}\|\hat{\beta}_j - \beta_j \|^2_{\ell_q}}
\label{eq:key}
\end{eqnarray}
Consider separately the cases where
\begin{eqnarray}\label{eq:inequality1}
3\sum_{j\in S(\beta)}\lambda_n{(d_j)^{ 1/q'}}\|\hat{\beta}_j -
\beta_j \|_{\ell_q} \leq  \frac{\delta}{n} \|X\beta - f^*
\|^2_{\ell_2}
\end{eqnarray}
and
\begin{eqnarray}\label{eq:inequality2}
3\sum_{j\in S(\beta)}\lambda_n{(d_j)^{ 1/q'}}\|\hat{\beta}_j -
\beta_j \|_{\ell_q} >  \frac{\delta}{n} \|X\beta - f^* \|^2_{\ell_2}
\end{eqnarray}
In case (\ref{eq:inequality1}), the result of the theorem trivially
follows from equation (\ref{eq:sparseoracle}). So ,we will only
consider the case (\ref{eq:inequality2}). All the subsequent
inequalities are valid on the event $\mathcal{A}\cap \mathcal{A}_1$
where $\mathcal{A}_1$ is defined by (\ref{eq:inequality2}). On this
event, we get from (\ref{eq:sparseoracle}) that
\begin{eqnarray}
\sum_{j=1}^{p_n}(d_j)^{1/q'}\|\hat{\beta}_j-\beta_j\|_{\ell_q} \leq
3\left(1 + \frac{1}{\delta} \right)\sum_{j\in S(\beta)}{(d_j)^{
1/q'}}\|\hat{\beta}_j - \beta_j \|_{\ell_q}
\end{eqnarray}
which further implies that
\begin{eqnarray}
\sum_{j \in
S(\beta)^c}(d_j)^{1/q'}\|\hat{\beta}_j-\beta_j\|_{\ell_q} \leq
\left(2 + \frac{3}{\delta} \right)\sum_{j\in S(\beta)}{(d_j)^{
1/q'}}\|\hat{\beta}_j - \beta_j \|_{\ell_q}
\end{eqnarray}
By assumption \ref{assumption.re}, we have
\begin{eqnarray}
\kappa(s',\delta)\sqrt{\sum_{j \in S(\beta)
}(d_j)^{2/q'-1}\|\hat{\beta}_j - \beta_j \|^2_{\ell_q}} &\leq&
\sqrt{\frac{1}{n}\|X(\hat{\beta}-\beta) \|^2_{\ell_2}}  =
\frac{1}{\sqrt{n}}\|\hat{Y} - X\beta\|_{\ell_2}
\end{eqnarray}
Combining this with (\ref{eq:key}), we get
\begin{eqnarray}
\frac{1}{n}\|\hat{Y} -f^* \|_{\ell_2}^2  & \leq &
\frac{1}{n}\|X\beta - f^*\|^2_{\ell_2} +
3\lambda_n\kappa^{-1}(s',\delta)\sqrt{\bar{d}_n|S(\beta)|}\left(
\frac{1}{\sqrt{n}}\|\hat{Y}
- X\beta \|_{\ell_2}\right) \\
& \leq &\frac{1}{n}\|X\beta - f^*\|^2_{\ell_2} +
4\lambda_n\kappa^{-1}(s',\delta)\sqrt{\bar{d}_n|S(\beta)|}\biggl(
\frac{1}{\sqrt{n}}\|\hat{Y} - f^* \|_{\ell_2}  \nonumber \\
& ~ & +\frac{1}{\sqrt{n}}\|X\beta - f^* \|_{\ell_2} \biggr)
\end{eqnarray}
This inequality is of the same form as (A.4) in
\citep{Bunea:annal:07}. A standard decoupling argument as in
\citep{Bunea:annal:07} using inequality $\ds 2xy \leq \frac{x^2}{b}
+ by^2$ with $b>1$, $x = \ds
\lambda_n\kappa^{-1}(s',\delta)\sqrt{\bar{d}_n|S(\beta)|}$, and $y$
being either $\ds \frac{1}{\sqrt{n}}\|\hat{Y} - f^* \|_{\ell_2}$ or
$\ds \frac{1}{\sqrt{n}}\|X\beta - f^* \|_{\ell_2}$ yields that
\begin{eqnarray}
\frac{1}{n}\|\hat{Y} - f^* \|_{\ell_2}^2 \leq
\frac{b+1}{b-1}\frac{1}{n}\|X\beta - f^*\|^2_{\ell_2} +
\frac{8b^2}{(b-1)\kappa^2(s',\delta)}\lambda_n^2\bar{d}_n|S(\beta)|,~~\forall
\beta
> 1.
\end{eqnarray}
Taking $b = 1 + 2/\delta$ in the last display finishes the proof of
the theorem. $\Box$

From the above sparse oracle inequalities,  we can show that the
$\ell_1\text{-}\ell_q$ regression estimator can achieve the optimal
rate of convergence if some ``{\it weak sparsity}" condition holds
\citep{Bunea:07}. The main intuition is, even if the true function
$f^*$ can not be represented exactly by a linear model $X\beta$, but
for some $\tilde{\beta} \in \mathbb{R}^{m_n}$ the squared distance
from $f^*$ to $X\beta$ can be controlled, up to logarithmic factors,
by $|S(\tilde{\beta})|/n$. Then, the optimal rate of convergence can
still be achieved. More formally, we define an oracle set as
\begin{definition}
Let $B$ be a constant depending only for $f^*$ and define an oracle
set as
\begin{eqnarray}
\mathcal{B} = \left\{\beta:~~\mathrm{s.t.}~~ \frac{1}{n}\|f^* -
X\beta \|_{\ell_2}^2 \leq B\lambda_n^2|S(\beta)| \right\}
\end{eqnarray}
\end{definition}

\begin{corollary}\label{corollary.rate1}
Under the same condition as in theorem \ref{thm.oracle}, if the
oracle set $\mathcal{B}$ is nonempty and there is at least one
element $\tilde{\beta}$ such that $|S(\tilde{\beta})|\leq s'$, we
have
\begin{eqnarray}
\frac{1}{n}\|f^* -\hat{Y}\|_{\ell_2}^2  = O_P\left( \frac{\bar{d}_n
s'\log m_n}{n} \right)
\end{eqnarray}
Therefore, when $s' \leq s_n$, the $\ell_1\text{-}\ell_q$ regression
estimator achieves the optimal rate of convergence.
\end{corollary}

\begin{remark}
Generally, the conditions for estimation consistency is weaker than
those for variable selection consistency. For $q=1$, why assumption
\ref{assumption.re} and \ref{assumption.re1} are weaker than the
assumptions in theorem \ref{thm.sparsistency} can be found in
 \citep{MY06} and \citep{Bick:Ritov:2007}. The cases for $q>1$ and
 the group cases should follow in a similar way.
\end{remark}

\section{Risk Consistency}\label{sec.persistency}

In this section, we study the risk consistency (or persistency)
property with random design, which holds under a much weaker
condition than variable selection consistency and does not need the
true model to be linear. Instead of directly to show the persistency
result for the estimator defined in equation \ref{eqn.grouplasso},
we show the persistency result for a constrained form estimator,
which is equivalent to the estimator in \ref{eqn.grouplasso} in the
sense of primal and dual problems.

Due to the fact of random design and increasing dimensions, the same
triangular array statistical paradigm as in \citep{GR04} is adopted.
In the following, we use calligraphic letter, such as $\mathcal{Z}$
to represent random variables, while $Z$ to represent its
realization. Consider the triangular array
$\mathcal{Z}_{1}^{(n)},\ldots, \mathcal{Z}_{n}^{(n)}$ (which is
simplified as $\mathcal{Z}_{1},\ldots \mathcal{Z}_{n}$), our study
mainly focus on the case where $\mathcal{Z}_1,\ldots,\mathcal{Z}_n \stackrel{iid}{\sim} F_n \in \mathcal{F}^n$,
where $\mathcal{F}^n$ is a collection of distributions
of $m_n + 1$ dimensional i.i.d. random vectors
\begin{eqnarray}
\mathcal{Z}_i = (\mathcal{Y}_i,
\mathcal{X}_{i,\underline{1}},\ldots, \mathcal{X}_{i,
\underline{m_n}})~~~i=1,\ldots, n
\end{eqnarray}
with the corresponding realizations
\begin{eqnarray}
Z_i =  (Y_i, X_{i,\underline{1}},\ldots, X_{i,
\underline{m_n}})~~~i=1,\ldots, n.
\end{eqnarray}

Denote
\begin{eqnarray}
\gamma = (-1,\beta_{\underline{1}},\ldots, \beta_{\underline{m_n}})
= (\beta_{\underline{0}},\beta_{\underline{1}},\ldots,
\beta_{\underline{m_n}}),
\end{eqnarray}
and define
\begin{eqnarray}
R_{F_n}(\beta) = \mathbb{E}\left(\mathcal{Y} -
\sum_{\underline{j}=1}^{m_n}\mathcal{X}_{\underline{j}}\beta_{\underline{j}}
\right)^2 = \gamma^T\Sigma_{F_n}\gamma
\end{eqnarray}
where $\mathcal{Z} =(\mathcal{Y}, \mathcal{X}_{\underline{1}},\ldots, \mathcal{X}_{\underline{m_n}}) \sim F_n \in \mathcal{F}^n$
and $(\Sigma_{F_n}) = \mathbb{E}\mathcal{Z}^T\mathcal{Z}$.

Given $n$ observations $Z_1,\ldots,Z_n$, denote their empirical distribution by $\hat{F}_n$
and define the empirical risk as
\begin{eqnarray}
R_{\hat{F}_n}(\beta) = \gamma^T\Sigma_{\hat{F}_n}\gamma
\end{eqnarray}
where $\ds \Sigma_{\hat{F}_n} = \frac{1}{n}\sum_{i=1}^n Z_iZ^T_i$.

Given a sequence of sets of predictors
$\mathcal{B}_n = \{ \sum_{j=1}^{p_n}(d_j)^{1/q'}\|\beta_j\|_{\ell_q}
\leq L_n \}$,
the sequence of estimators
$\hat{\beta}^{\hat{F}_n}$ is called persistent if for every sequence $F_n \in \mathcal{F}^n$,
\begin{eqnarray}
R_{F_n}(\hat{\beta}^{\hat{F}_n}) - R_{F_n}(\beta_{*}^{F_n})
\stackrel{P}{\rightarrow} 0,
\end{eqnarray}
where
\begin{eqnarray}
\hat{\beta}^{\hat{F}_n} & = & \argmin_{\beta \in \mathcal{B}_n} R_{\hat{F}_n}(\beta) = \argmin_{\beta \in \mathcal{B}_n} \|Y-X\beta\|_{\ell_2}^2\\
\beta_{*}^{F_n} & = & \argmin_{\beta \in \mathcal{B}_n} R_{F_n}(\beta).
\end{eqnarray}

To show the persistency result, a moment condition as in
\citep{Zhou:07} is needed.

\begin{assumption}\label{assumption.moments}
For each $\underline{j},\underline{k} \in \{1,\ldots, m_n+1\}$,
denote $E = (\mathcal{Z}\mathcal{Z}^T - \mathbb{E}(\mathcal{Z}\mathcal{Z}^T))_{\underline{j},\underline{k}}$,
where $\mathcal{Z} = (\mathcal{Y},
\mathcal{X}_{\underline{1}},\ldots, \mathcal{X}_{
\underline{m_n}})$, suppose that there exists some constants $M$ and
$s$.
\begin{eqnarray}
\mathbb{E}(|E|^q) \leq
q!M^{q-2}s/2
\end{eqnarray}
for every $q \geq 2$ and every $F_n \in \mathcal{F}^n$.
\end{assumption}

\begin{theorem}\label{thm.persistency}
Suppose that $\ds m_n \leq e^{n^\xi}$ for some $\xi < 1$. If $\ds
L_n = o\left((n/\log n)^{1/4}\right)$, then $\ell_1\text{-}\ell_q$
regularized regression is persistent. That is, for every sequence
$F_n \in \mathcal{F}^n$:
\begin{eqnarray}
R_{F_n}(\hat{\beta}^{\hat{F}_n}) - R_{F_n}(\beta_*^{F_n}) = o_P(1).
\end{eqnarray}
\end{theorem}

\noindent{\bf Proof:}  For any $\underline{j},\underline{k} \in
\{1,\ldots, m_n+1\}$ and any $\delta > 0$,  from assumption
\ref{assumption.moments} we can apply the Bernstein's inequality and
obtain
\begin{eqnarray}
\ds \mathbb{P}\Bigl(
\bigl|\left({\Sigma_{\hat{F}_n}}\right)_{\underline{j},\underline{k}}
- \left(\Sigma_{F_n}\right)_{\underline{j},\underline{k}}\bigr|
> \delta\Bigr) \leq e^{-cn\delta^2}
\end{eqnarray}
for some $c > 0$. Therefore, by Bonferoni bound we have
\begin{eqnarray}
{ \ds \mathbb{P}\Bigl(
\max_{\underline{j},\underline{k}}\bigl|\left({\Sigma_{\hat{F}_n}}\right)_{\underline{j},\underline{k}}
- \left(\Sigma_{F_n}\right)_{\underline{j},\underline{k}}\bigr|
> \delta\Bigr)  \leq m^2_ne^{-cn\delta^2} \leq
e^{2n^\xi-cn\delta^2} \leq e^{-cn\delta^2/2}}
\end{eqnarray}
for large enough $n$.  For a sequence $\ds\delta_n =
\sqrt{\frac{2\log n}{cn}}$, we have
\begin{eqnarray}
 \mathbb{P}\Bigl(
\max_{\underline{j},\underline{k}}\bigl|\left({\Sigma_{\hat{F}_n}}\right)_{\underline{j},\underline{k}}
- \left(\Sigma_{F_n}\right)_{\underline{j},\underline{k}}\bigr|
> \delta_n\Bigr) \leq \frac{1}{n} \rightarrow 0
\end{eqnarray}
which implies that
\begin{eqnarray}
\max_{\underline{j},\underline{k}}\bigl|\left({\Sigma_{\hat{F}_n}}\right)_{\underline{j},\underline{k}}
- \left(\Sigma_{F_n}\right)_{\underline{j},\underline{k}}\bigr| =
O_P\left(\sqrt{\frac{\log n}{n}}\right).
\end{eqnarray}
Therefore,
\begin{eqnarray}
\sup_{\beta \in \mathcal{B}_n}\bigl|R_{F_n}(\beta) - R_{\hat{F}_n}(\beta) \bigr| & = &
\sup_{\beta \in \mathcal{B}_n}\bigl|\gamma^T(\Sigma_{F_n} - \Sigma_{\hat{F}_n})\gamma \bigr| \\
& \leq &
\max_{\underline{j},\underline{k}}\bigl|\left({\Sigma_{\hat{F}_n}}\right)_{\underline{j},\underline{k}}
- \left(\Sigma_{F_n}\right)_{\underline{j},\underline{k}}\bigr| \|
\gamma\|^2_{\ell_1}
\\
& \leq &
\max_{\underline{j},\underline{k}}\bigl|\left({\Sigma_{\hat{F}_n}}\right)_{\underline{j},\underline{k}}
-
\left(\Sigma_{F_n}\right)_{\underline{j},\underline{k}}\bigr|\left(1
+ \sum_{j=1}^{p_n}\| \beta_j\|_{\ell_1}\right)^2
\\
& \leq &
\max_{\underline{j},\underline{k}}\bigl|\left({\Sigma_{\hat{F}_n}}\right)_{\underline{j},\underline{k}}
-
\left(\Sigma_{F_n}\right)_{\underline{j},\underline{k}}\bigr|\left(1
+ \sum_{j=1}^{p_n}(d_j)^{1/q'}\| \beta_j\|_{\ell_q}\right)^2
\\
&\leq&
\max_{\underline{j},\underline{k}}\bigl|\left({\Sigma_{\hat{F}_n}}\right)_{\underline{j},\underline{k}}
-
\left(\Sigma_{F_n}\right)_{\underline{j},\underline{k}}\bigr|(1+L_n)^2
=o_P(1)\nonumber
\end{eqnarray}
for $\ds L_n = o\left((n/\log n)^{1/4}\right)$.

Further, by definition, we have
$R_{\hat{F}_n}(\hat{\beta}^{\hat{F}_n})\leq R_{F_n}(\beta_*^{F_n})$,
combining with the following inequalities
\begin{eqnarray}
R_{F_n}(\hat{\beta}^{\hat{F}_n})
-{R}_{\hat{F}_n}(\hat{\beta}^{\hat{F}_n}) &\leq & \sup_{\beta \in
{\mathcal{B}_n}} \bigl|R_{F_n}(\beta) - {R}_{\hat{F}_n}(\beta)
\bigr| \\
{R}_{\hat{F}_n}(\beta^{\hat{F}_n}_*)
-{R}_{F_n}({\beta}^{\hat{F}_n}_*) &\leq & \sup_{\beta \in
{\mathcal{B}_n}} \bigl|R_{F_n}(\beta) - {R}_{\hat{F}_n}(\beta)
\bigr|.
\end{eqnarray}
This implies that
\begin{eqnarray}
R_{F_n}(\hat{\beta}^{\hat{F}_n}) - R_{F_n}(\beta_{*}^{F_n}) \leq 2
\sup_{\beta \in {\mathcal{B}_n}} \bigl|R_{F_n}(\beta) -
{R}_{\hat{F}_n}(\beta) \bigr| = o_P(1),
\end{eqnarray}
which completes the proof. $\Box$

\section{Discussions}\label{sec.conclusion}

The results presented here show that many good properties from
$\ell_1\text{-}$regularization (Lasso) naturally carry on to the
$\ell_1\text{-}\ell_q$ cases ($1 \leq q \leq \infty$), even if the
number of variables within each group also increase with the sample
size $n$. Using fixed design, we get both variable selection and
estimation consistency under different conditions. Using random
design, we get persistency under a much weaker condition. Our
results provide a unified treatment for both the iCAP estimator
($q=\infty$) and the group Lasso estimator ($q=2$).

Our results can also provide theoretical analysis to the
simultaneous Lasso estimator \citep{Turlach05, Tropp06c} for joint
sparsity. Which can find a good approximation of several response
variables at once using different linear combinations of the high
dimensional covariates. At the same time, it tries to balance the
error in approximation against the total number of covariates that
participate. Assuming that we have altogether $\bar{d}_n$ response,
the $i$-th signal is represented as $Y^{(i)} \in \mathbb{R}^n$, and
the design matrix is $X=(X_{\underline{1}},\ldots,
X_{\underline{p_n}}) \in \mathbb{R}^{n\times p_n}$. Denote the model
as
\begin{eqnarray}
Y^{(i)} = X\beta^{(i)} + \epsilon^{(i)},~~~i=1,\ldots, \bar{d}_n
\end{eqnarray}
The simultaneous Lasso estimator can be formulated as
\begin{equation}\label{eqn.simlasso}
\hat{\beta}^{(1)},\ldots, \hat{\beta}^{(\bar{d}_n)} =
\mathop{\argmin}_{\beta^{(1)},\ldots, \beta^{(\bar{d}_n)}}
\frac{1}{2n}\sum_{k=1}^{\bar{d}_n}\left\|Y^{(k)} -
X\beta^{(k)}\right\|^2_{\ell_2} + \lambda_n
\sum_{j=1}^{p_n}\max_{\underline{\ell} \in \{1,\ldots, \bar{d}_n
\}}|\beta^{(j)}_{\underline{\ell}}|,
\end{equation}
This problem can be formulated as a standard $\ell_1\text{-}\ell_q$
regularized regression estimator with $q=\infty$. For this, define
\begin{eqnarray}
\tilde{Y} = \left(
              \begin{array}{c}
                Y^{(1)} \\
                \vdots \\
                Y^{(\bar{d}_n)}\\
              \end{array}
            \right)\in
            \mathbb{R}^{n\bar{d}_n}~~~\tilde{X}=I_{\bar{d}_n}\otimes
            X = \left(\begin{array}{ccc}
                  X &  &  \\
                   & \ddots &  \\
                   &  & X
                \end{array}\right)~~\mathrm{and}~\beta= \left(
              \begin{array}{c}
                \beta^{(1)} \\
                \vdots \\
                \beta^{(\bar{d}_n)}\\
              \end{array}
            \right)
\end{eqnarray}
where $\otimes$ denotes the Kronecker product. Therefore, the
simultaneous Lasso estimator can be rewritten as
\begin{equation}
\hat{\beta}^{(1)},\ldots, \hat{\beta}^{(\bar{d}_n)} =
\mathop{\argmin}_{\beta^{(1)},\ldots, \beta^{(\bar{d}_n)}}
\frac{1}{2n}\left\|\tilde{Y} - \tilde{X}\beta\right\|^2_{\ell_2} +
\lambda'_n \sum_{j=1}^{p_n}(\bar{d}_n)\max_{\underline{\ell} \in
\{1,\ldots, \bar{d}_n \}}|\beta^{(j)}_{\underline{\ell}}|
\end{equation}
where $\lambda'_n = \lambda_n/\bar{d}_n$.  This is just an
$\ell_1\text{-}\ell_\infty$ regularized regression estimator with
block design. Therefore, all results in this paper can be applied to
analyze such type estimators.

\section{Acknowledgements}

We thank John Lafferty,   Pradeep Ravikumar, Alessandro Rinaldo,
 Larry Wasserman, and
Shuheng Zhou for their very helpful discussions and comments.

\bibliography{paper}
\bibliographystyle{ims}

\end{document}